\documentclass{article}

\usepackage[dblblindworkshop,final]{neurips_2026}
\workshoptitle{Agentic AI for Biological Discovery (AgenticLS)}

\usepackage[utf8]{inputenc}
\usepackage[T1]{fontenc}
\usepackage{hyperref}
\hypersetup{hidelinks}
\usepackage{url}
\usepackage{graphicx}
\usepackage{booktabs}
\usepackage{amsmath}
\usepackage{amsfonts}
\usepackage{amssymb}
\usepackage{microtype}
\usepackage{courier}
\usepackage{etoc}
\usepackage{adjustbox}

\title{Auditing Tool Traces in Agentic LLMs: Source Localization and Counterfactual Dependence Testing}
\author{%
  \textbf{Junhao Liu\textsuperscript{1}\quad Tianxiao Li\textsuperscript{2}\quad Yongxian Wu\textsuperscript{1}\quad Ziheng Duan\textsuperscript{1}}\\
  \textbf{Ren Li\textsuperscript{1}\quad Jing Zhang\textsuperscript{1}\quad Martin Renqiang Min\textsuperscript{2}}\\[0.5em]
  \textsuperscript{1}University of California, Irvine\\
  \textsuperscript{2}NEC Labs America
}

\begin{document}
\etocdepthtag.toc{mainmatter}

\maketitle
\begin{abstract}
    Agentic LLMs increasingly support scientific prediction by grounding decisions in external evidence. In biology, they promise accurate, interpretable predictions for important problems by integrating complex experimental and computational evidence. However, serialized tool results can contain hidden label shortcuts, such as dataset identifiers or template artifacts, that create apparent success without biological reasoning. To address this risk, we introduce a two-stage audit. Stage~1, \textbf{Per-Tool TF-IDF Decomposition (PT2D)}, localizes label-predictive patterns to individual tool results while avoiding signal dilution in the concatenated tool trace. Stage~2, \textbf{Token-Class Redaction}, counterfactually tests whether the frozen LLM depends on the localized shortcut. We apply this audit to TCR--pMHC binding, where tool augmentation substantially outperforms a no-evidence baseline. At face value, this suggests that the LLM has learned to use scientific evidence. The audit overturns this interpretation: PT2D localizes the strongest signal to identifiers in one tool result, and redacting them reduces performance to chance. These findings show how hidden label shortcuts can masquerade as scientific reasoning or discovery, motivating source-level audits before benchmark gains are trusted.
\end{abstract}

\section{Introduction}

Large language models (LLMs) are increasingly coupled to external evidence in scientific workflows. General paradigms include tool use \citep{schick2023toolformer,yao2023react}, retrieval-augmented generation \citep{lewis2020rag}, and agentic pipelines that chain multiple tools and return their results to the model \citep{shen2023hugginggpt}; domain systems have instantiated these ideas with genomics databases and chemistry tools \citep{jin2024genegpt,bran2024chemcrow}. We use \emph{tool trace} to denote the ordered collection of serialized tool results supplied to the LLM. In scientific workflows, these results are intended to serve as external evidence for prediction, but the resulting tool trace remains distinct from a reasoning trace generated by the LLM itself.

The appeal of this setup in biomedical settings is straightforward. Parametric knowledge can be incomplete or stale, and LLMs remain prone to hallucination \citep{lewis2020rag,jin2024genegpt,bran2024chemcrow}. Injecting structured evidence from databases, simulators, or analytical tools at inference time therefore appears to give the model a chance to reason from retrieved facts rather than unsupported recall, while keeping predictions auditable against an explicit tool trace.

However, evidence-augmented pipelines introduce an evaluation failure mode that ordinary benchmark checks may miss \citep{kapoor2023leakage,lapuschkin2019cleverhans}. Serialized tool results may contain hidden label shortcuts that correlate with labels without carrying the intended scientific semantics. Such shortcuts \citep{geirhos2020shortcut} can arise from entry identifiers, version strings, file paths, duplicated fields, template artifacts, or other textual patterns introduced by deposition, annotation, sampling, or serialization. They may survive semantic data splits because their generating process need not follow scientific grouping. Screening only the concatenated tool trace can also miss a pattern confined to one component because heterogeneous results dilute it. Thus, apparently rigorous performance may reflect shortcut exploitation rather than evidence-grounded reasoning.


To address hidden label shortcuts in tool traces, we introduce a two-stage audit that separates source localization from behavioral verification. First, \textbf{Per-Tool TF-IDF Decomposition (PT2D)} analyzes each tool result independently to localize label-predictive patterns that may be hidden when heterogeneous results are concatenated into a full tool trace. This design is motivated by the \emph{dilution effect}: unrelated components can obscure a shortcut confined to one tool. Second, \textbf{Token-Class Redaction} removes the localized shortcut class from all prompt-bearing fields and reevaluates the same frozen LLM, counterfactually testing whether its predictions depend on that shortcut rather than the intended scientific evidence. We further validate the motivation for per-tool decomposition by quantifying dilution through real tool-subset sweeps and synthetic label-independent additions.

In our TCR--pMHC case study using peptide-disjoint splits from prior benchmarks \citep{slone2025stag,abeer2026unseen}, a Qwen3-4B model \citep{yang2025qwen3} fine-tuned on five-tool traces substantially outperforms a no-evidence baseline, seemingly indicating evidence-grounded generalization. The audit overturns this interpretation: PT2D localizes the strongest transferable signal to PDB identifiers in one tool result, and identifier redaction collapses performance to chance while peptide redaction leaves it unchanged. Real-tool and label-independent synthetic dilution sweeps further explain why full-trace screening misses this shortcut. Together, these findings motivate source-level provenance analysis and counterfactual verification before tool-augmented gains are interpreted as scientific reasoning or discovery.

\section{Method: Two-Stage Shortcut Audit}

Let $\mathcal{D}_{\mathrm{fit}}$ and $\mathcal{D}_{\mathrm{eval}}$ be labeled audit partitions aligned with the target evaluation protocol; the audit is otherwise split-agnostic. Each example is $(x_i,y_i,E_i)$, where $x_i$ is the input, $y_i\in\{0,1\}$ is the label, and $E_i=[e_i^{(1)},\ldots,e_i^{(T)}]$ is a tool trace comprising one serialized result $e_i^{(t)}$ from each of $T$ tools. We target label-predictive patterns in $e_i^{(t)}$ that lack intended task semantics but correlate with $y_i$ because of dataset construction, identifier assignment, or serialization.

Our running example is binary TCR-pMHC binding: a downstream LLM predicts binding from a candidate TCR, its peptide--MHC context, and evidence from five structural-biology tools (Appendix~\ref{app:five-tool-example}). The audit separates source localization from model-dependence testing: Stage~1, Per-Tool TF-IDF Decomposition (PT2D), localizes label-predictive patterns to individual tool results; Stage~2, Token-Class Redaction, counterfactually tests whether the frozen LLM behaviorally depends on the localized shortcut.

\begin{figure*}[t]
    \centering
    \begin{minipage}[t]{0.37\textwidth}
        \centering
        \includegraphics[width=\linewidth]{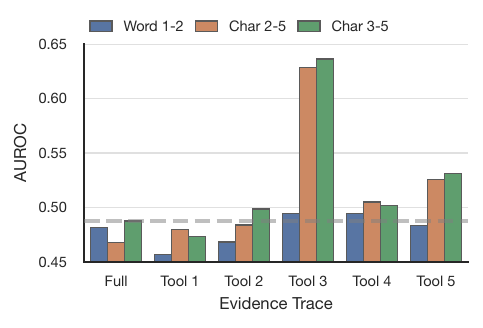}
        \caption{PT2D localizes Tool~3 as the strongest cross-split carrier.}
        \label{fig:pt2d-localization}
    \end{minipage}
    \hfill
    \begin{minipage}[t]{0.60\textwidth}
        \centering
        \includegraphics[width=\linewidth]{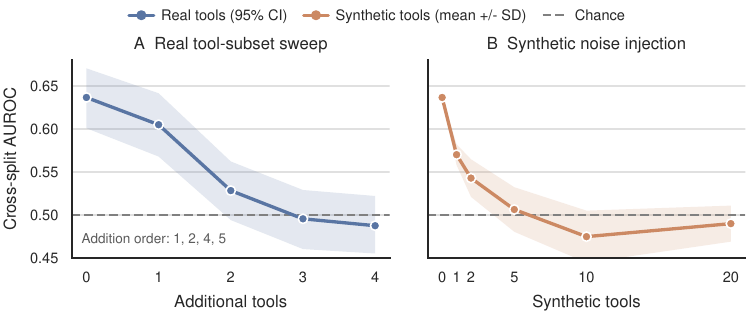}
        \caption{Adding real (A) or synthetic (B) tool results dilutes Tool~3's cross-split signal.}
        \label{fig:dilution-effect}
    \end{minipage}
\end{figure*}

\paragraph{Per-Tool TF-IDF Decomposition (PT2D)}

PT2D addresses a potential \emph{dilution effect}: concatenating unrelated tool results can obscure a shortcut confined to one component and cause a classifier trained on the full tool trace to miss it. PT2D therefore audits each tool section separately and retains the concatenated trace only for comparison.

For each scope $s\in\{\mathrm{full},1,\ldots,T\}$, we construct TF-IDF features $\phi_s$ \citep{salton1988termweighting}, fit balanced logistic regression $g_s$ on $\mathcal{D}_{\mathrm{fit}}$, and score $\mathcal{D}_{\mathrm{eval}}$. Let $E_i^{(\mathrm{full})}$ concatenate all sections, $E_i^{(t)}=e_i^{(t)}$, and $\hat y_i^{(s)}=g_s(\phi_s(E_i^{(s)}))$. We report
$A_s=\mathrm{AUROC}\!\left(\{y_i\}_{i\in\mathcal{D}_{\mathrm{eval}}},
    \{\hat y_i^{(s)}\}_{i\in\mathcal{D}_{\mathrm{eval}}}\right)$.    

Word $n$-grams capture lexical content, whereas character $n$-grams expose identifier fragments. We rank carriers by comparing each $A_t$ with the full-trace scope, chance, and other tools, without a universal threshold.
PT2D returns a suspect source and discriminative substrings without LLM retraining or parameter access. Its evidence is associational: a predictive text pattern need not be used by the downstream LLM. Appendix~\ref{app:pt2d} gives complete settings and feature-provenance analysis.

\paragraph{Token-Class Redaction}

We trace PT2D's suspect source and substrings to serialized fields, grouping values with the same generating semantics into candidate token classes such as entry identifiers, version strings, or file-path components. We target the complete class because an isolated TF-IDF $n$-gram may cover only part of a shortcut-bearing value; recurring, non-overlapping classes serve as controls. PT2D thus proposes what to test but does not establish downstream LLM usage.

Following input-erasure analyses \citep{li2016erasure}, we redact each candidate class separately. For class $c$, $R_c$ replaces every occurrence in all prompt-bearing fields with a class-specific placeholder while preserving labels and reference answers. We evaluate the same frozen LLM with identical generation, parsing, and scoring and report $\Delta A_c=A_{\mathrm{orig}}-A_{R_c}$, where positive values denote performance lost. Global replacement prevents duplicated values from surviving elsewhere; testing one class at a time at a frozen checkpoint isolates the intervention from retraining.

We interpret a large performance drop after targeted redaction, coupled with stable performance under a control redaction, as evidence that the model relies on the localized shortcut. This conclusion is limited to model behavior, because redaction may also remove correlated signals or shift the input distribution.

\section{Experiments}


\paragraph{Task and data.}
TCR--pMHC binding provides a biologically grounded setting for testing whether LLMs can interpret amino-acid sequences with domain-specific computational evidence. Our case study asks whether an LLM can combine its parametric knowledge with results from protein-analysis tools to classify binding. Concretely, the model predicts a binary label from paired TCR$\alpha$/TCR$\beta$ CDR3 sequences, a peptide, its MHC allele and sequence, and a five-tool trace. We use the dataset from \citet{slone2025stag} and the peptide-disjoint cluster splits of \citet{abeer2026unseen}. Cluster~0 supplies 12,940 training (3,483 positive) and 1,438 validation (397 positive) examples; clusters~1--5 contain 4,898 test examples (264 positive).

\paragraph{Tool uses.}
Each tool trace has five fixed-order tool results: (1) CDR3$\beta$ biophysical descriptors (length, composition, charge, flexibility, and contact potential); (2) McPAS-TCR retrieval and similarity \citep{tickotsky2017mcpas}; (3) ThermoMPNN mutation-level $\Delta\Delta G$ \citep{dieckhaus2024thermompnn}; (4) CDR3$\beta$ positional stability and hotspots derived from (3), adapted from Biomni \citep{huang2025biomni}; and (5) structure-based CDR3$\beta$--pMHC interface descriptors (buried surface area computed with FreeSASA \citep{mitternacht2016freesasa}, contact distance, hydrogen bonds, and salt bridges). Tools receive candidate fields and the modeled complex, never labels (Appendix~\ref{app:five-tool-example} as example).

\subsection{PT2D Localizes Leakage}
\label{sec:pt2d-results}

\paragraph{Model and evaluation.}
We perform supervised fine-tuning of Qwen3-4B \citep{yang2025qwen3} on examples that pair the task fields and complete five-tool trace with the corresponding binding label. Unless otherwise stated, we compute test AUROC jointly over all 4,898 examples from clusters~1--5; for redaction, we additionally report AUROC within each cluster. The resulting model achieves $A_{\mathrm{orig}}=0.8136$ test AUROC on the combined clusters~1--5 test set. For context, a label-balanced no-evidence model reaches only $0.4928$ test AUROC in our development comparison. Appendix~\ref{app:experimental-details} reports training and evaluation details.
Before interpreting $A_{\mathrm{orig}}=0.8136$ as evidence-grounded biological generalization, we apply PT2D to determine whether tool results contain a transferable textual shortcut, which tool carries it, and what serialized field produces it.
For this case protocol, $\mathcal{D}_{\mathrm{fit}}$ is cluster~0 and $\mathcal{D}_{\mathrm{eval}}$ is peptide-disjoint clusters~1--5. We compute $A_s$ for the full tool trace and five tool sections using word $1$--$2$-grams, character $2$--$5$-grams, and character $3$--$5$-grams, testing transfer to the same held-out clusters as the LLM evaluation.
Figure~\ref{fig:pt2d-localization} reports $A_s$ under the cross-split PT2D setting described above. At the full-trace scope, all three feature sets remain near chance ($A_{\mathrm{full}}=0.4818$, $0.4677$, and $0.4876$), so screening only the concatenated tool results would suggest little transferable leakage. Per-tool decomposition reverses this conclusion: Tool~3 (\texttt{predict-thermompnn-ddg}) reaches $A_3=0.6288$ and $0.6365$ with character features, while Tool~5, which also serializes the PDB identifier, shows a weaker increase to $A_5=0.5259$--$0.5310$; Tools~1, 2, and~4 remain near chance. For character $3$--$5$-grams, the gap $A_3-A_{\mathrm{full}}=0.1489$ shows that aggregation hides a strong local signal. Because $A_s$ is produced by the diagnostic logistic regression rather than the downstream LLM, this result establishes an exploitable shortcut in the tool results but does not yet establish LLM reliance.

Logistic-regression weights localize the signal from Tool~3 to PDB-ID fragments in \texttt{ThermoMPNN\_inference\_<pdb\_id>.csv}, rather than biological language: word features remain near chance ($A_3=0.4946$), whereas character features reach $A_3=0.6365$, with prefixes such as \texttt{\_21} favoring binders and \texttt{\_17} favoring non-binders. A two-digit PDB-prefix lookup reaches AUROC $0.6967$, motivating complete PDB-ID redaction in the next stage; Appendix~\ref{app:pt2d} reports the full analysis.
We next ask why full-trace screening misses the Tool~3 leakage localized above. Starting with Tool~3 and progressively appending Tools~1, 2, 4, and~5 causes the character $3$--$5$-gram diagnostic AUROC to decrease monotonically from $0.6365$ to $0.4876$; label-independent synthetic additions reproduce this decay, reaching chance by $K=5$ (Figure~\ref{fig:dilution-effect}). This dilution explains the full-trace false negative and motivates per-tool decomposition; Appendix~\ref{app:dilution} reports the complete analysis.

\begin{table}[t]
    \centering
    \small
    \begin{tabular}{lrr}
        \toprule
        Variant          & AUROC $\uparrow$ & $\Delta A_c$ \\
        \midrule
        Original         & 0.8136           & ---          \\
        PDB-ID redacted  & 0.4972           & $+0.3163$    \\
        Peptide redacted & 0.8152           & $-0.0017$    \\
        \bottomrule
    \end{tabular}
    \caption{Frozen-checkpoint AUROC after targeted token-class redaction ($\Delta A_c=A_{\mathrm{orig}}-A_{R_c}$).}
    \label{tab:token-redaction}
\end{table}

\subsection{Redaction Confirms Identifier Dependence}

PT2D establishes that PDB identifiers carry a transferable label signal, but not whether the fine-tuned LLM actually uses it. We therefore perform a frozen-model counterfactual: we remove every PDB identifier from all prompt-bearing fields and reevaluate the same checkpoint on the same examples, while separately redacting peptide sequences as a non-overlapping control. A selective performance drop after PDB-ID redaction would provide behavioral evidence that the model depends on the localized shortcut; Appendix~\ref{app:redaction} reports the intervention details and robustness results.
Removing PDB IDs reduces AUROC from $A_{\mathrm{orig}}=0.8136$ to $A_{R_{\mathrm{PDB}}}=0.4972$, giving $\Delta A_{\mathrm{PDB}}=0.3163$ and eliminating nearly all measured above-chance ranking performance (Table~\ref{tab:token-redaction}).
Peptide redaction yields $A_{R_{\mathrm{peptide}}}=0.8152$ and $\Delta A_{\mathrm{peptide}}=-0.0017$, showing no loss and ruling out direct reuse of the literal epitope sequence. Together, PT2D localization and the frozen-model counterfactual provide evidence that behavior in this TCR task depends on dataset-specific identifiers rather than only the intended biological evidence. The redaction supports dependence on the complete identifier class; short-prefix attribution comes only from PT2D weights.

\section{Conclusion}

Tool-augmented scientific LLMs can appear to reason from domain evidence while relying on hidden label shortcuts embedded in serialized tool results. We introduce a two-stage audit that separates source localization from behavioral verification: PT2D exposes predictive patterns that full-trace screening can miss through dilution, while Token-Class Redaction tests whether a frozen model actually depends on the localized shortcut. In the TCR--pMHC case study, this audit overturned an apparent generalization success by tracing it to PDB identifiers and showing that removing the identifier class eliminated nearly all above-chance performance while a non-overlapping control left performance intact. We hope this work draws greater attention to the need for per-source provenance analysis and counterfactual verification of what a model actually uses before its behavior is interpreted as scientific reasoning.



\bibliographystyle{plainnat}
\bibliography{main}

\cleardoublepage
\appendix

  \noindent{\Large\bfseries Appendix\par}
  \vspace{0.4em}

\etocdepthtag.toc{appendix}
\begingroup
  \etocsettagdepth{mainmatter}{none}
  \etocsettagdepth{appendix}{subsection}
  \renewcommand{\contentsname}{Contents}
  \tableofcontents
\endgroup

\begin{center}
  \rule{\linewidth}{1pt}
\end{center}

\section{Limitations}

Our experiments are conducted on a single TCR-pMHC benchmark using one 4B backbone and one tool suite; the precise dilution behavior may therefore vary with the length, format, and content of the evidence in other settings. Although PT2D is partition-agnostic, its fitting and evaluation partitions must be aligned with the evaluation protocol of the target application. At the methodological level, the audit is post-hoc: it detects rather than prevents shortcuts, and its conclusions concern observable dependence at the input--output level rather than the internal mechanisms through which that dependence arises.

\section{Experimental Details}
\label{app:experimental-details}

\subsection{Data and splits}
\label{app:data-splits}

Following the peptide-disjoint protocol of \citet{abeer2026unseen}, the split unit is the peptide cluster rather than an individual TCR--peptide pair. Cluster~0 supplies the training and validation examples, while clusters~1--5 form the peptide-disjoint test set. Table~\ref{tab:appendix-split-counts} reports the class counts used by both the LLM evaluation and the cross-split text diagnostics. The underlying data remain imbalanced; label balancing refers to replacement sampling during SFT and to class-weighted logistic regression in PT2D, not to discarding negative examples.

\begin{table}[!ht]
    \centering
    \small
    \begin{tabular}{lrrr}
        \toprule
        Split           & Total  & Positive & Negative \\
        \midrule
        Train (C0)      & 12,940 & 3,483    & 9,457    \\
        Validation (C0) & 1,438  & 397      & 1,041    \\
        Test C1         & 1,004  & 45       & 959      \\
        Test C2         & 994    & 38       & 956      \\
        Test C3         & 968    & 30       & 938      \\
        Test C4         & 968    & 109      & 859      \\
        Test C5         & 964    & 42       & 922      \\
        \midrule
        Test (C1--C5)   & 4,898  & 264      & 4,634    \\
        \bottomrule
    \end{tabular}
    \caption{Dataset composition. Test clusters C1--C5 contain peptides disjoint from cluster~0.}
    \label{tab:appendix-split-counts}
\end{table}

\subsection{Tool trace and prompt structure}
\label{app:prompt-structure}

Each prompt contains the TCR$\alpha$/TCR$\beta$ CDR3 sequences, peptide, MHC allele and sequence, followed by five fixed-order tool-result sections. Tool~3 serializes its source filename as \texttt{ThermoMPNN\_inference\_<pdb\_id>.csv}; Tool~5 separately emits the same value as \texttt{complex\_id}. This duplication motivates applying redaction to every prompt-bearing field rather than editing only the visibly suspicious section.

\begin{table*}[!t]
    \centering
    \small
    \begin{tabular}{clp{0.59\textwidth}}
        \toprule
        Step & Tool                             & Tool-result content                                                                                        \\
        \midrule
        1    & CDR3$\beta$ biophysical analysis & Length, residue composition, charge, flexibility, and contact-potential descriptors.                       \\
        2    & McPAS-TCR retrieval              & Database matches, sequence similarity, and associated epitope annotations.                                 \\
        3    & ThermoMPNN $\Delta\Delta G$      & Mutation-level stability predictions and the serialized inference filename.                                \\
        4    & Stability summary                & Positional stability and hotspot summaries derived from Tool~3.                                            \\
        5    & TCR--pMHC interface analysis     & Buried surface area, contact distances, hydrogen bonds, salt bridges, and a serialized complex identifier. \\
        \bottomrule
    \end{tabular}
    \caption{Fixed tool-trace order used during SFT and evaluation.}
    \label{tab:appendix-tools}
\end{table*}

\subsection{Training and generation evaluation}
\label{app:training-evaluation}

The label-balanced Qwen3-4B model \citep{yang2025qwen3} is trained for 10 epochs with rank-16 LoRA adapters \citep{hu2022lora} ($\alpha=32$) on all linear layers, learning rate $10^{-5}$, and global batch size 32. The formal redaction study evaluates the final update-4040 checkpoint. All counterfactual variants use this same frozen checkpoint, the same 4,898 test rows, no assistant prefill, and identical decoding and parsing.

The model generates an integer $a\in\{1,\ldots,10\}$ inside \texttt{<answer>}; evaluation maps it to $s=(a-1)/9$ before computing AUROC. Format validity requires a parseable answer in this range. The original, PDB-ID-redacted, peptide-redacted, and format-redacted conditions are fully valid.

\section{Full PT2D Results and Identifier Provenance}
\label{app:pt2d}

\subsection{Classifier specification}
\label{app:pt2d-specification}

For the TCR instantiation, PT2D is fit on cluster~0 and evaluated directly on clusters~1--5. Word features use \texttt{TfidfVectorizer(analyzer=word, ngram\_range=(1,2), max\_features=30000, min\_df=3)}. Character $2$--$5$-grams use \texttt{analyzer=char\_wb} with 60,000 maximum features, and character $3$--$5$-grams use 50,000 maximum features; both use \texttt{min\_df=3}. Every scope uses logistic regression with $C=1$, the \texttt{liblinear} solver, balanced class weights, and 4,000 maximum iterations.

\begin{table}[t]
    \centering
    \small
    \begin{tabular}{lrrr}
        \toprule
        Scope      & Word 1--2 & Char 2--5       & Char 3--5       \\
        \midrule
        Full trace & 0.4818    & 0.4677          & 0.4876          \\
        Tool 1     & 0.4569    & 0.4797          & 0.4733          \\
        Tool 2     & 0.4681    & 0.4839          & 0.4985          \\
        Tool 3     & 0.4946    & \textbf{0.6288} & \textbf{0.6365} \\
        Tool 4     & 0.4947    & 0.5050          & 0.5015          \\
        Tool 5     & 0.4833    & 0.5259          & 0.5310          \\
        \bottomrule
    \end{tabular}
    \caption{PT2D AUROC computed jointly over all examples from test clusters C1--C5. Character features isolate Tool~3 despite the near-chance full-trace diagnostic.}
    \label{tab:appendix-pt2d-full}
\end{table}

\begin{table*}[t]
    \centering
    \small
    \begin{tabular}{lrrrrrr}
        \toprule
        Scope      & All             & C1     & C2     & C3     & C4              & C5     \\
        \midrule
        Full trace & 0.4876          & 0.4888 & 0.5413 & 0.6359 & 0.4565          & 0.4216 \\
        Tool 1     & 0.4733          & 0.3769 & 0.4687 & 0.5548 & 0.4881          & 0.4403 \\
        Tool 2     & 0.4985          & 0.4997 & 0.5957 & 0.5708 & 0.4706          & 0.4160 \\
        Tool 3     & \textbf{0.6365} & 0.5566 & 0.6392 & 0.6109 & \textbf{0.6999} & 0.4874 \\
        Tool 4     & 0.5015          & 0.4510 & 0.5369 & 0.5727 & 0.4878          & 0.5037 \\
        Tool 5     & 0.5310          & 0.4982 & 0.5018 & 0.5445 & 0.5702          & 0.4717 \\
        \bottomrule
    \end{tabular}
    \caption{Character $3$--$5$-gram PT2D AUROC by peptide-disjoint test cluster. All denotes AUROC computed jointly over C1--C5.}
    \label{tab:appendix-pt2d-clusters}
\end{table*}

\subsection{Feature provenance and prefix lookup}
\label{app:identifier-provenance}

The highest-weight Tool~3 character features point to a single serialization template. Positive-weight fragments include \texttt{nce\_8}, \texttt{\_21}, \texttt{ce\_21}, \texttt{\_16}, and \texttt{ce\_16}; negative-weight fragments include \texttt{\_17}, \texttt{ce\_17}, \texttt{\_22}, \texttt{ce\_9}, \texttt{\_18}, and \texttt{\_23}. These strings occur at the boundary between \texttt{inference\_} and the PDB identifier, rather than in biological prose or predicted $\Delta\Delta G$ values.

As a field-level check, we learn a prefix-to-binder-rate lookup on cluster~0 and use the learned rate as the test score. Table~\ref{tab:appendix-prefix-lookup} shows that two digits yield 0.6967 AUROC when evaluated jointly over C1--C5. This establishes that short prefixes are transferable label correlates; it does not by itself show that the LLM relies only on those two digits. The complete-ID intervention in Appendix~\ref{app:redaction} supplies the behavioral evidence at the identifier-class level.

\begin{table*}[t]
    \centering
    \small
    \begin{tabular}{lrrrrrr}
        \toprule
        Prefix   & All             & C1     & C2     & C3     & C4     & C5     \\
        \midrule
        1 digit  & 0.6725          & 0.5776 & 0.5686 & 0.6542 & 0.7470 & 0.5202 \\
        2 digits & \textbf{0.6967} & 0.5794 & 0.7416 & 0.5946 & 0.7716 & 0.4967 \\
        3 digits & 0.6793          & 0.4820 & 0.6522 & 0.5614 & 0.8442 & 0.5817 \\
        \bottomrule
    \end{tabular}
    \caption{Cross-split AUROC of an explicit PDB-prefix lookup learned on cluster~0. All denotes AUROC computed jointly over C1--C5.}
    \label{tab:appendix-prefix-lookup}
\end{table*}

\section{Token-Class Redaction Details and Robustness}
\label{app:redaction}

\subsection{Replacement rules}
\label{app:redaction-rules}

For every test row, redaction is applied to \texttt{evidence}, \texttt{question}, their duplicates in \texttt{extra\_info}, and every message content field. PDB-ID redaction replaces the numeric suffix in \texttt{ThermoMPNN\_inference\_<id>.csv}, the value of \texttt{complex\_id}, and any remaining exact occurrence of the row's identifier with \texttt{<PDB\_ID>}. Peptide redaction replaces every exact epitope sequence with \texttt{<PEPTIDE>}. Labels and reference answers are never modified.

Format redaction is a broader sensitivity check that replaces step headers, trace banners, section headers, XML tags, and field names with \texttt{<FORMAT>}. Table~\ref{tab:appendix-redaction-counts} quantifies the intervention sizes.

\begin{table}[t]
    \centering
    \small
    \begin{tabular}{lrl}
        \toprule
        Variant & Replacements & Placeholder        \\
        \midrule
        PDB ID  & 50,778       & \texttt{<PDB\_ID>} \\
        Peptide & 63,674       & \texttt{<PEPTIDE>} \\
        Format  & 3,582,603    & \texttt{<FORMAT>}  \\
        \bottomrule
    \end{tabular}
    \caption{Total replacements across the 4,898 test prompts and their duplicated prompt-bearing fields.}
    \label{tab:appendix-redaction-counts}
\end{table}

\subsection{Complete generation results}
\label{app:redaction-results}

Table~\ref{tab:appendix-redaction-full} reports the targeted redaction conditions and the format sensitivity analysis. PDB-ID and peptide redaction are the targeted intervention and non-overlapping control used for the main claim. Format redaction preserves output validity but removes multiple overlapping structural cues; its decrease indicates sensitivity to trace scaffolding, not an independent token-class attribution.

\begin{table*}[t]
    \centering
    \small
    \begin{tabular}{lrrrrrrr}
        \toprule
        Variant          & All    & $\Delta A_c$ & C1     & C2     & C3     & C4     & C5     \\
        \midrule
        Original         & 0.8136 & ---          & 0.5746 & 0.8298 & 0.8324 & 0.8686 & 0.8322 \\
        PDB-ID redacted  & 0.4972 & $+0.3163$    & 0.5126 & 0.5241 & 0.4818 & 0.4963 & 0.4858 \\
        Peptide redacted & 0.8152 & $-0.0017$    & 0.5763 & 0.8535 & 0.8437 & 0.8646 & 0.8295 \\
        Format redacted  & 0.5189 & $+0.2947$    & 0.4787 & 0.4829 & 0.4343 & 0.5747 & 0.5184 \\
        \bottomrule
    \end{tabular}
    \caption{Formal frozen-checkpoint generation AUROC. All denotes AUROC computed jointly over C1--C5. Positive $\Delta A_c=A_{\mathrm{orig}}-A_{R_c}$ denotes performance lost after redaction. The broad format intervention is a sensitivity analysis rather than a clean control.}
    \label{tab:appendix-redaction-full}
\end{table*}

\section{Dilution Experiments: Complete Protocol and Results}
\label{app:dilution}

\subsection{Real-tool subset sweep}
\label{app:real-dilution}

The real sweep fixes Tool~3 as the carrier and adds Tools~1, 2, 4, and 5 in that fixed order. Every point refits the same character $3$--$5$-gram PT2D classifier. Confidence intervals are percentile intervals from 2,000 bootstrap resamples stratified by label within the combined test set (clusters~1--5). The endpoints reproduce the Tool~3-only and full-trace values exactly.

\begin{table}[t]
    \centering
    \small
    \begin{tabular}{lrrl}
        \toprule
        Tools         & $|S|$ & AUROC  & 95\% CI          \\
        \midrule
        \{3\}         & 1     & 0.6365 & [0.6008, 0.6704] \\
        \{3,1\}       & 2     & 0.6049 & [0.5679, 0.6415] \\
        \{3,1,2\}     & 3     & 0.5284 & [0.4941, 0.5622] \\
        \{3,1,2,4\}   & 4     & 0.4956 & [0.4605, 0.5291] \\
        \{3,1,2,4,5\} & 5     & 0.4876 & [0.4553, 0.5221] \\
        \bottomrule
    \end{tabular}
    \caption{Real-tool dilution sweep with Tool~3 fixed as the carrier.}
    \label{tab:appendix-real-dilution}
\end{table}

\subsection{Label-independent synthetic additions}
\label{app:synthetic-dilution}

The synthetic source pool contains 53,512 Tool~1, 2, and 4 sections from train and test, with mean length 1,845.9 characters. For each target example and synthetic slot, the procedure independently samples a source section and a target length, then repeats or clips the source text to that length. Sampling never uses the target label or cluster, and train and test use independent random-number streams. We evaluate seeds 20260517--20260521 for each $K\in\{0,1,2,5,10,20\}$.

\begin{table}[t]
    \centering
    \small
    \begin{tabular}{rrrrr}
        \toprule
        $K$ & Mean   & SD     & Min    & Max    \\
        \midrule
        0   & 0.6365 & 0.0000 & 0.6365 & 0.6365 \\
        1   & 0.5701 & 0.0121 & 0.5590 & 0.5891 \\
        2   & 0.5429 & 0.0218 & 0.5256 & 0.5809 \\
        5   & 0.5064 & 0.0259 & 0.4821 & 0.5489 \\
        10  & 0.4749 & 0.0302 & 0.4464 & 0.5249 \\
        20  & 0.4900 & 0.0209 & 0.4669 & 0.5167 \\
        \bottomrule
    \end{tabular}
    \caption{AUROC is computed jointly over test clusters C1--C5 and summarized over five synthetic-dilution seeds. The curve reaches chance by five added sections and remains near chance thereafter.}
    \label{tab:appendix-synthetic-dilution}
\end{table}

\section{Real Five-Tool Trace}
\label{app:five-tool-example}

To make the serialized tool trace concrete, we show one real trace in abridged form from the cluster~4 test example: TCR$\alpha$ CDR3 \texttt{CAEGGGSYIPTF}, TCR$\beta$ CDR3 \texttt{CASSEGDTEAFF}, peptide \texttt{KLVALGINAV}, PDB ID \texttt{21009}, and label~1. The fields and values below are copied from the stored tool results. To fit the page, we preserve representative tool-result fields and the decision signal from every tool while omitting the full MHC sequence, repeated preview and position rows, local filesystem prefixes, and other redundant fields; no displayed value is recomputed.

\paragraph{Tool 1: CDR3$\beta$ biophysics.}
\begin{flushleft}
    \small\ttfamily
    candidate\_cdr3b: CASSEGDTEAFF\\
    length: 12; charge\_net: -3\\
    hydrophobic\_count\_pct: 5/41.7\\
    polar\_count\_pct: 4/33.3\\
    negative\_count\_pct: 3/25.0\\
    central\_region\_sequence: EGDT\\
    BIOPHYSICS\_SIGNAL:\\
    charge=NEGATIVE\_BIAS;\\
    flexibility=BALANCED\_FLEXIBILITY;\\
    contact=MODERATE\_CONTACT\_POTENTIAL
\end{flushleft}

\paragraph{Tool 2: McPAS-TCR retrieval.}
\begin{flushleft}
    \small\ttfamily
    total\_entries: 40731\\
    exact\_epitope\_hits: 50\\
    known\_cdr3b\_count\_for\_exact\_epitope: 44\\
    most\_similar\_known\_cdr3b: CASSAGGDTEAFF\\
    most\_similar\_sequence\_identity: 0.8800\\
    DATABASE\_SUPPORT\_SIGNAL:\\
    STRONG\_DATABASE\_SUPPORT
\end{flushleft}

\paragraph{Tool 3: ThermoMPNN mutation-level predictions.}
\begin{flushleft}
    \small\ttfamily
    thermompnn\_csv\_path:\\
    .../ThermoMPNN\_inference\_21009.csv\\
    total\_predictions: 2160\\
    preview\_ddg\_mean: 1.815 kcal/mol\\
    preview\_ddg\_min: 0.000 kcal/mol\\
    preview\_ddg\_max: 2.708 kcal/mol\\
    THERMOMPNN\_PREVIEW\_SIGNAL:\\
    PREVIEW\_SHOWS\_DESTABILIZING\_RISK
\end{flushleft}

\paragraph{Tool 4: CDR3$\beta$ stability summary.}
\begin{flushleft}
    \small\ttfamily
    predictions\_used: 2160\\
    mapped\_sequence\_check: PASS\\
    avg\_position\_sensitivity\_ddg: 0.427 kcal/mol\\
    counts\_tolerant\_moderate\_sensitive: 7/4/1\\
    hotspot: F126 (mean\_ddg=1.761)\\
    STRUCTURAL\_STABILITY\_SIGNAL:\\
    STRUCTURALLY\_STABLE
\end{flushleft}

\paragraph{Tool 5: CDR3$\beta$--pMHC interface.}
\begin{flushleft}
    \small\ttfamily
    complex\_id: 21009\\
    cdr3\_total\_bsa\_A2: 123.92\\
    cdr3\_polar\_bsa\_A2: 71.53\\
    cdr3\_apolar\_bsa\_A2: 52.39\\
    min\_contact\_distance\_A: 2.75\\
    num\_hydrogen\_bonds: 0\\
    num\_salt\_bridges: 0\\
    contacting\_residues: 3/19\\
    hotspots: D124(58.53), G108(37.86), T125(27.51)\\
    STRUCTURE\_INTERFACE\_SIGNAL:\\
    PARTIAL\_CDR3\_PMHC\_ENGAGEMENT
\end{flushleft}

\section{Information About Use of AI Assistants}
\label{app:ai-assistance}

AI assistants were used to support the writing and revision of the manuscript, the preparation of figures, and the development of experimental code. The authors carefully proofread and revised all AI-assisted text and reviewed and verified all AI-assisted figures and code. The authors take full responsibility for the accuracy and integrity of the manuscript and all associated research artifacts.

\end{document}